\documentclass[letterpaper, 10 pt, conference]{ieeeconf}

\IEEEoverridecommandlockouts
\usepackage[english]{babel}
\usepackage{subfigure}
\usepackage{bm}
\usepackage{subfig}
\usepackage{amsmath}
\usepackage{tensor}
\usepackage{amsfonts}
\usepackage{amssymb}
\usepackage{graphicx}
\usepackage{url}
\usepackage{breqn}
\usepackage{siunitx}
\usepackage{afterpage}
\usepackage{subfigure}
\usepackage{multirow}
\usepackage{here} 
\usepackage{soul}
\usepackage{tabularx}
\newcolumntype{b}{X}
\newcolumntype{s}{>{\hsize=.3\hsize}X}
\newcolumntype{x}{>{\hsize=.5\hsize}X}
\usepackage{tikz}
\usetikzlibrary{shapes.geometric, arrows}

\usepackage{array}

\newcommand{\camf}{\mathcal{C}}
\newcommand{\robotf}{\mathcal{B}}
\newcommand{\worldf}{\mathcal{I}}

\newcommand{\vece}{\mathbf{e}}

\newcommand{\vecn}{\mathbf{n}}

\newcommand{\vecxi}{\bm{\xi}}

\newcommand{\vecp}{\mathbf{p}}
\newcommand{\vecq}{\mathbf{q}}

\newcommand{\vecu}{\mathbf{u}}

\newcommand{\vecx}{\mathbf{x}}

\newcommand{\mat}[2]{\mathbf{#1}_{#2}}

\newcommand{\matK}{\mathbf{K}}

\newcommand{\matU}{\mathbf{U}}
\newcommand{\matR}{\bm{\mathit{R}}}
\newcommand{\matX}{\mathbf{X}}

\newcommand{\inertia}{\mathbf{J}}

\newcommand{\loadmass}{m_{L}}

\newcommand{\massQ}{m_{Q}}
\newcommand{\half}{\frac{1}{2}}

\newcommand{\angvel}{\mathbf{\Omega}}

\newcommand{\robotpos}[1]{\vecx_{#1}}
\newcommand{\robotrot}[1]{\matR_{#1}}

\newcommand{\robotvel}[1]{\dot{\vecx}_{#1}}
\newcommand{\robotacc}[1]{\ddot{\vecx}_{#1}}
\newcommand{\robotangvel}[1]{\angvel_{#1}}

\newcommand{\loadpos}{\vecx_{L}}
\newcommand{\loadposdes}{\vecx_{L,des}}

\newcommand{\loadvel}{\dot{\vecx}_{L}}

\newcommand{\loadacc}{\ddot{\vecx}_{L}}

\newcommand{\cablevec}[1]{\vecxi_{#1}}

\newcommand{\cabledotvec}[1]{\dot{\vecxi}_{#1}}

\newcommand{\quat}{\vecq}
\newcommand{\quatdot}{\mathbf{\dot{q}}}

\newcommand{\inputforce}{\vecu}

\newcommand{\aptpos}[1]{\vecp_{#1}}
\newcommand{\aptvel}[1]{\dot{\vecp}_{#1}}

\newcommand{\realnum}[1]{\mathbb{R}^{#1}}
\newcommand{\SOthree}{SO(3)}

\newcommand{\norm}[1]{\left\lVert#1\right\rVert}
\newcommand{\abs}[1]{\left\lvert#1\right\rvert}
\newcommand{\twonorm}[1]{\left\lVert#1\right\rVert_2}
\newcommand{\prths}[1]{\left(#1\right)}
\newcommand{\brcks}[1]{\left[#1\right]}

\newcommand{\qcamframe}[1]{\camf_{#1}}
\newcommand{\axis}[2]{\mathbf{e}_{#1}^{#2}}

\usepackage{cite}

\usepackage{wasysym}  
\author{Guanrui Li$^*$, Alex Tunchez$^*$, and Giuseppe Loianno
\thanks{$^*$These authors contributed equally.}
\thanks{The authors are with the New York University, Tandon School of Engineering, Brooklyn, NY 11201, USA. {\tt\footnotesize email: \{lguanrui, atunchez, loiannog\}@nyu.edu}.}
\thanks{This work was supported by Qualcomm Research, the Technology Innovation Institute, Nokia, NYU Wireless, and the young researchers program "Rita Levi di Montalcini" 2017 grant PGR17W9W4N.}
}

\title{\LARGE \bf PCMPC: Perception-Constrained Model Predictive Control for Quadrotors with Suspended Loads using a Single Camera and IMU}

\begin{document}

\maketitle
\thispagestyle{empty}
\pagestyle{empty}

\begin{abstract}
In this paper, we address the Perception--Constrained Model Predictive Control (PCMPC) and state estimation problems for quadrotors with cable suspended payloads using a single camera and Inertial Measurement Unit (IMU). We design a receding--horizon control strategy for cable suspended payloads directly formulated on the system manifold configuration space SE(3)$\times$S$^2$. The approach considers the system dynamics, actuator limits and the camera's Field Of View (FOV) constraint to guarantee the payload's visibility during motion. The monocular camera, IMU, and vehicle's motor speeds are combined to provide estimation of the vehicle's states in 3D space, the payload's states, the cable's direction and velocity. The proposed control and state estimation solution runs in real-time at 500 Hz on a small quadrotor equipped with a limited computational unit. The approach is validated through experimental results considering a cable suspended payload trajectory tracking problem at different speeds.
\end{abstract}

\IEEEpeerreviewmaketitle
\section*{Supplementary material}
Video: \url{https://youtu.be/BBWt1xG7Rrw}
\section{Introduction}
Quadrotor Unmanned Aerial Vehicles (UAVs) can be employed in a variety of real-world applications such as inspection, maintenance, search and rescue, and package delivery. The versatility and mobility of UAVs has boosted the need to create autonomous aerial robots interacting with the environment, like transporting and manipulating objects. Aerial transportation can offer a faster and more versatile solution compared to ground transportation in congested urban environments and in post-disaster scenarios where aerial robots can deliver supplies. 

A typical aerial transportation setup uses either an active or passive manipulator like a gripper attached to the aerial vehicle~\cite{ThLoPoSrVk2014, garimella2015towards, kim2013aerial, heredia2014control} or a cable suspended from the vehicle~\cite{cruz2017cable, palunko2012agile, foehn2017fast, sreenath2013trajectory, dai2014adaptive, bernard2009generic}. However, a cable mechanism generally presents reduced inertia compared to a gripper. Such an advantage potentially leads to more agile motions and quicker rejection of perturbations. Moreover, active grasping mechanisms are harder to design and more energy-consuming compared to passive ones. However, cable suspended systems are under--actuated leading to challenges in control, planning and estimation. In this work, we investigate the  Perception--Constrained Model Predictive Control (PCMPC) and state estimation problems for a quadrotor with a cable--suspended payload using a single camera and IMU. 
\begin{figure}[!t]
  \centering
    \includegraphics[width=0.9\columnwidth]{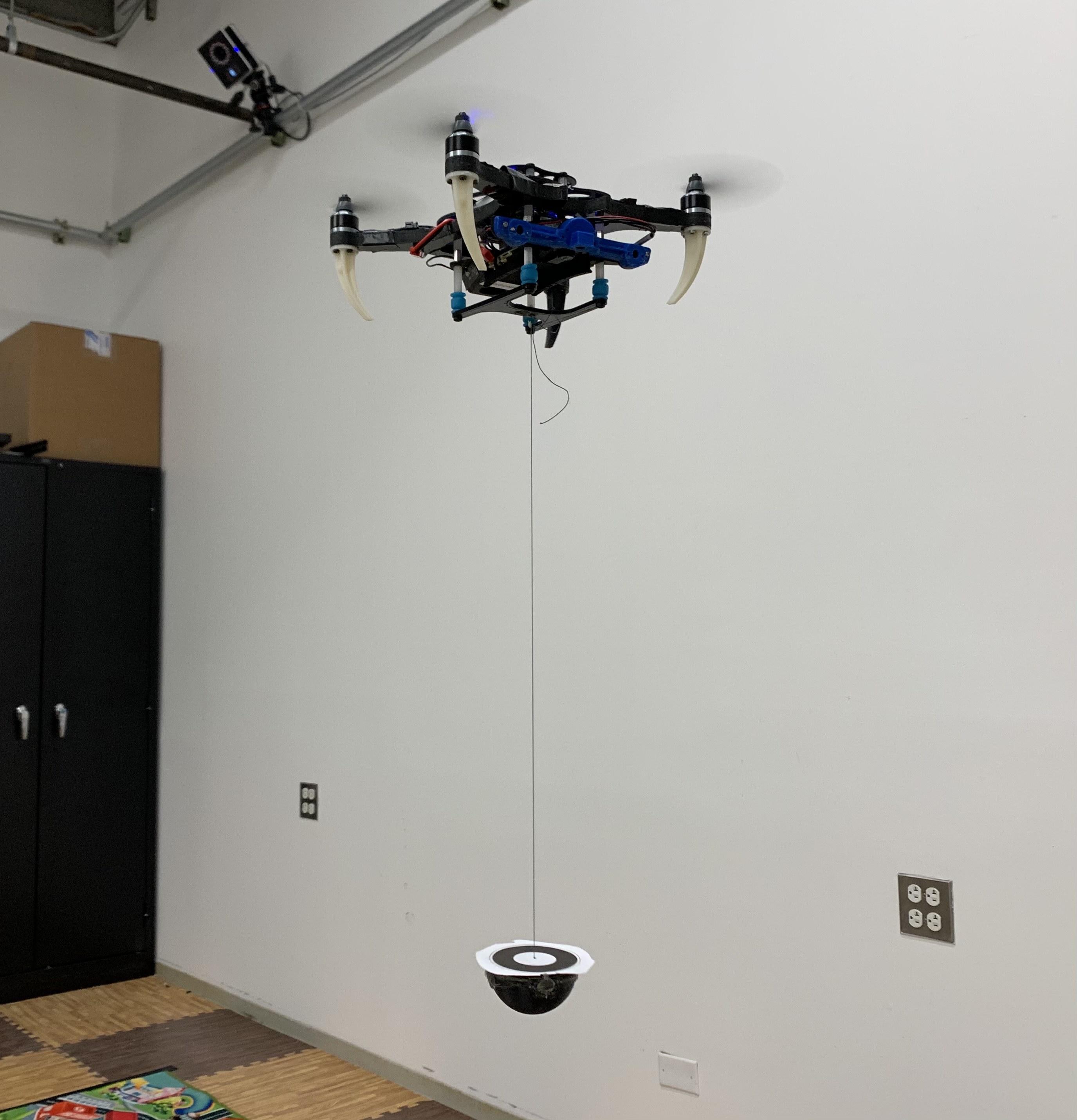}
  \caption{The quadrotor carrying a cable--suspended payload using PCMPC with monocular vision and IMU. \label{fig:Inro}}
  \vspace{-10pt}
\end{figure}
 \begin{figure*}[!t]
   \centering
     \includegraphics[width=\linewidth]{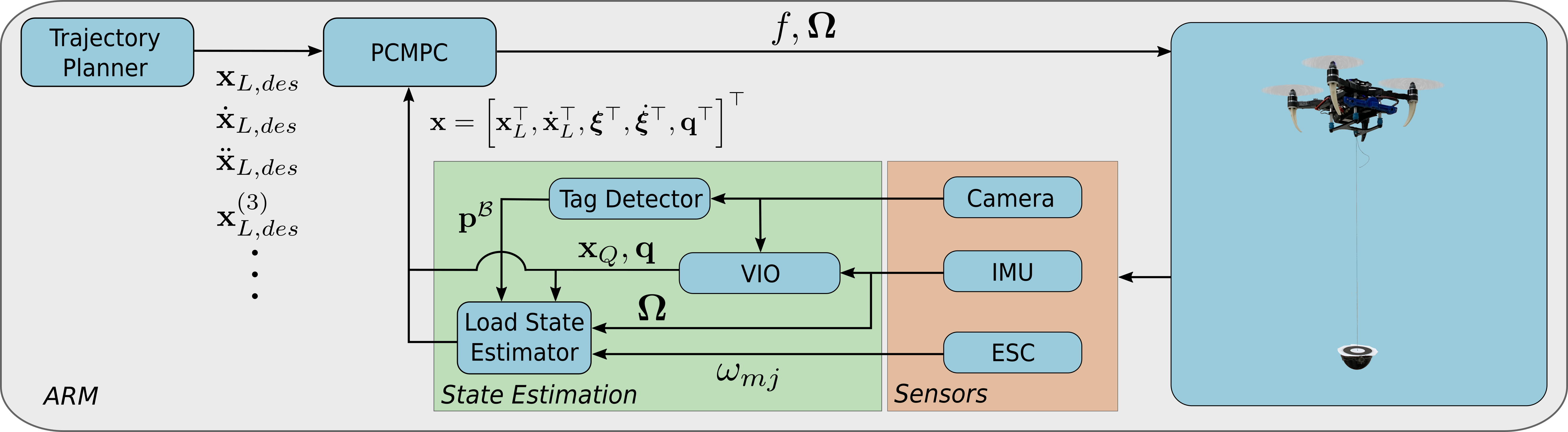}
   \caption{System architecture.\label{fig:BlockDiagram}}
   \vspace{-10pt}
 \end{figure*}

Several past works on cable--suspended payload transportation using aerial vehicles have focused on control and planning methods minimizing payload swings~\cite{palunko2012trajectory, guerrero2017swing}, including reinforcement learning or neural networks based control~\cite{Aleksandra2017, faust-quad-icra-13, palunko2012agile}. These methods restrict the full--range motion of a quadrotor with a cable--suspended payload. Moreover, they lead to sub--optimal energy performance and prevent the execution of agile motion since they use simplified system dynamics. Nonlinear and geometric controls such as \cite{sreenath2013geometric, goodarzi2015geometric} consider the full nonlinear dynamics of the system for the trajectory tracking of the payload. Finally, in~\cite{SonICRA2018} a MPC for a quadrotor with cable--suspended payload is presented. 
However, all aforementioned works did not address the perception–based
control and state estimation problems for cable suspended
payloads. They either rely on external motion capture systems or present only simulation results.

Recently, \cite{tang2018aggressive} employs a nonlinear geometric controller with an on--board camera detecting the payload for trajectory tracking of the suspended payload. However, the work does not resolve the state estimation since it still depends on the motion capture system to localize the vehicle. Furthermore, this approach and all previous contributions do not address the perception--based control problem enforcing system dynamics, actuators and sensing constraints during motion. In our proposed work, this aspect is addressed considering a MPC approach that enforces the payload to lie in the camera's FOV, respects the actuator constraints and considers the full nonlinear system dynamics. The state estimation problem is resolved by inferring the the vehicle’s states in  3D  space,  the  payload’s  states,  the  cable’s  direction and velocity using a single camera, IMU, and motor speeds.

This work presents multiple contributions. First, we present a Perception--Constrained Model Predictive Control for a quadrotor with a cable--suspended payload on the system manifold configuration space $SE(3)\times S^2$. Our proposed control respects the full nonlinear system dynamics, actuator limits, and payload visibility in the quadrotor's camera. Second, the vehicle's rotor speeds, monocular camera, and IMU  are used to estimate the vehicle's pose, cable's direction and velocity, as well as the payload's position and velocity. Third, our entire pipeline runs on--board a small quadrotor in real time at $500$ Hz. Finally, this is the first quadrotor with a cable--suspended payload using on--board perception--constrained control and state estimation which respects the full nonlinear system dynamics and actuator constraints. We believe that the proposed system can be deployed in real-world settings such as warehouses and GPS-denied environments and is applicable to a wide range of Micro Aerial Vehicles (MAVs).


\section{Modeling, Control, and Planning}\label{sec:modeling}
We introduce the dynamic model, PCMPC, and trajectory planning for a quadrotor with a cable--suspended payload. The system overview is presented in Fig.~\ref{fig:BlockDiagram}. The relevant variables used in our paper are stated in Table~\ref{tab:notation}. 
\subsection{System Dynamics}
\begin{figure}[!t]
  \centering
    \includegraphics[width=0.8\columnwidth]{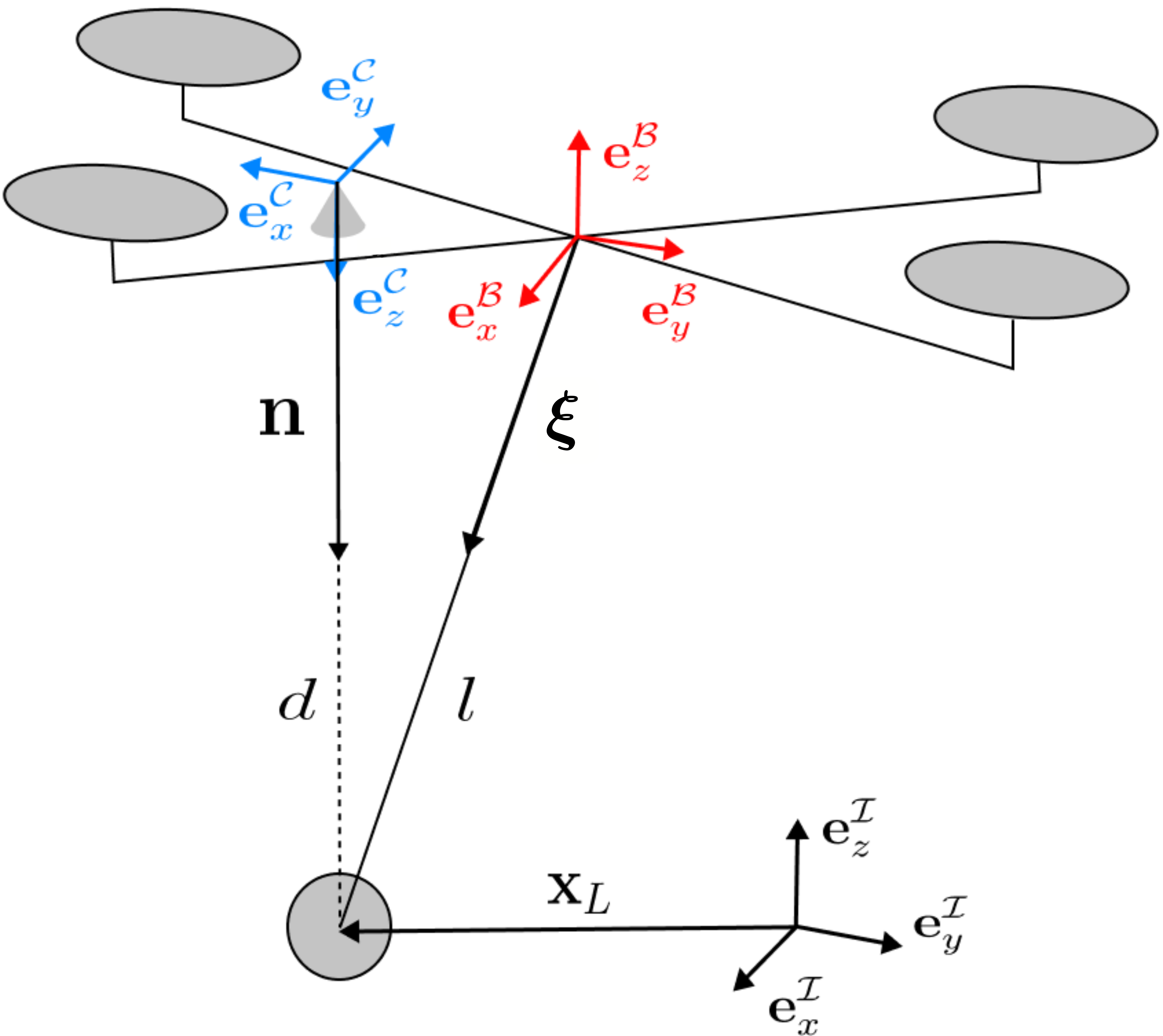}
  \caption{System  convention with $\mathcal{I}$, $\mathcal{C}$, and $\mathcal{B}$ denoting inertial, camera, and body frames, respectively.
  \label{fig:system_overview}}
  \vspace{-20pt}
\end{figure}

In Fig.~\ref{fig:system_overview} the physical configuration of the quadrotor and cable--suspended payload is depicted. It illustrates three different coordinate frames with axes \{$\vece{}^{\mathcal{I}}_x,\vece{}^{\mathcal{I}}_y,\vece{}^{\mathcal{I}}_z$\}, \{$\vece{}^{\mathcal{B}}_x,\vece{}^{\mathcal{B}}_y,\vece{}^{\mathcal{B}}_z$\}, 
\{$\vece{}^{\mathcal{C}}_x,\vece{}^{\mathcal{C}}_y,\vece{}^{\mathcal{C}}_z$\}, for the inertial, body, and camera frames, respectively. The goal is to derive the dynamic model in the $\mathcal{I}$ frame to formulate our PCMPC. The system dynamics are based on~\cite{sreenath2013trajectory}. The kinetic energy, $\mathcal{T}$, and potential energy, $\mathcal{U}$, of the system are defined as
\begin{align}
    \mathcal{T} &=\half\loadmass\twonorm{\loadvel}^2+\half\massQ\twonorm{\loadvel-l\cabledotvec{}}^2
    +\frac{1}{2}\angvel{}\cdot\inertia\angvel{},\label{eq:KinEng}\\
    \mathcal{U} &= (\massQ+\loadmass)g\loadpos{}\cdot\vece{}^{\mathcal{I}}_z - \massQ gl\cablevec{}\cdot\vece{}^{\mathcal{I}}_z.\label{eq:PotEng}
\end{align}

\begin{table}[!t]
\caption {Notation table.\label{tab:notation}} 
\centering
\begin{tabularx}{0.48\textwidth}{>{\hsize=0.58\hsize}X >{\hsize=1.42\hsize}X}
    \hline\hline
 $\worldf$, $\camf$, $\robotf$ & inertial, camera, robot frame\\
 $\loadmass,\massQ$ &  mass of payload, robot\\
 $\loadpos,\robotpos{Q}\in\realnum{3}$ &  position of payload, robot in $\worldf$\\
 $\loadvel, \loadacc\in\realnum{3}$ & linear velocity, acceleration of payload in $\worldf$\\
$\robotvel{Q}, \robotacc{Q}\in\realnum{3}$ &linear velocity, acceleration of robot in $\worldf$\\
 $\robotrot{}\in\SOthree$&orientation of robot with respect to $\worldf$ \\
  $\inertia_{}\in\realnum{3\times3}$ & moment of inertia robot\\
  $\robotangvel{}\in\realnum{3}$& angular velocity of robot in $\robotf$\\
 $\cablevec{} \in S^2$&unit vector from robot to attach point\\
$f\in\realnum{}$&total thrust\\
 $l,T\in\realnum{}$& length, tension of cable\\
 $g\in\realnum{}$& gravity constant\\
    \hline\hline
\end{tabularx}
\vspace{-1em}
\end{table}

Following the Lagrange d'Alembert principle, similar to~\cite{sreenath2013trajectory}, we obtain the system dynamics
\begin{align}
    \frac{d\loadpos}{dt} &= \loadvel\label{eq:acc},\\
    \prths{\massQ+\loadmass}\prths{\loadacc{}+g\vece{}^{\mathcal{I}}_z} &= \prths{\cablevec{}\cdot f\robotrot{}\vece{}^{\mathcal{I}}_z - m_Ql\twonorm{\cabledotvec{}}^2}\cablevec{},\label{eq:load_acc}\\
    \frac{d\cablevec{}}{dt}  &= \cabledotvec{},
        \end{align}

    \begin{align}
    m_Ql\ddot{\bm{\xi}}+m_Ql\twonorm{\cabledotvec{}}^2\cablevec{}&=\cablevec{}\times\prths{\cablevec{}\times f\robotrot{}\vece{}^{\mathcal{I}}_z},\label{eq:cable_acc}\\
    \quatdot{} &= \frac{1}{2}\Lambda\left(\angvel{}\right)\cdot \quat{}. \label{eq:quatdot}
\end{align}
The system dynamics $\mathbf{f}(\vecx{},\vecu{})$ considers $\vecx{} = [\loadpos^{\top}, \loadvel^{\top}, \cablevec{}^{\top},  \cabledotvec{}^{\top},\quat{}^{\top}]^\top$ and $\vecu = [f, \angvel{}^{\top}]^\top$ as the state and input vectors, respectively.
The state vector includes the quaternion $\quat{} = \left[q_w,q_x,q_y,q_z\right]^\top$ to describe the orientation $\robotrot{}$ of the quadrotor in the $\mathcal{I}$ frame. We employ this orientation parameterization instead of a rotation matrix so that we can reduce our state space size and consequently speed up our PCMPC as described in Section~\ref{sec:mpc}. The time derivative of the quaternion is expressed in eq.~(\ref{eq:quatdot}), where $\Lambda(\angvel{})$ is a skew-symmetric matrix of the quadrotor angular velocity, $\angvel = \left[\Omega_x, \Omega_y,\Omega_z\right]^\top$. This is shown in the following equation
\begin{equation}
   \Lambda(\angvel{}) = 
   \begin{bmatrix}
     0 & -\Omega_x & -\Omega_y & -\Omega_z \\
     \Omega_x & 0 & \Omega_z & -\Omega_y \\
     \Omega_y & -\Omega_z & 0 & \Omega_x \\
     \Omega_z & \Omega_y & -\Omega_x & 0
   \end{bmatrix}.
\end{equation}

The total thrust is the lift generated by each of the rotors $f=f_1+f_2+f_3+f_4$ along $\vece{}^{\mathcal{B}}_z$. Subsequently, it is necessary to apply a rotation induced by the quaternion onto $\vece{}^{\mathcal{I}}_z$. This is represented in eq.~(\ref{eq:load_acc}) by using an equivalent operation
\begin{equation}
    \quat{}\odot\vece{}^{\mathcal{I}}_z = \robotrot{}\vece{}^{\mathcal{I}}_z, \label{eq:rotmat}
\end{equation}
where
\begin{equation*}
    \robotrot{} =  
    \begin{bmatrix}
     1-2\prths{q_y^2+q_z^2} & 2\prths{q_xq_y-q_wq_z} & 2\prths{q_wq_y+q_xq_z} \\
     2\prths{q_wq_z+q_xq_y} & 1-2\prths{q_x^2+q_z^2} & 2\prths{q_yq_z-q_wq_x} \\
     2\prths{q_xq_z -q_wq_y} & 2\prths{q_wq_x+q_yq_z} & 1-2\prths{q_x^2+q_y^2}\\
   \end{bmatrix}.
\end{equation*}

\subsection{Perception--Constrained Model Predictive Control}~\label{sec:mpc}
Conventional control implementations, like PID as in \cite{tang2018aggressive} and \cite{KS:TL:13}, are not able to optimize a cost for a certain trajectory. Additionally, they do not guarantee boundaries on the inputs are satisfied. On the other hand, a MPC approach leverages the system dynamics under applied constraints to optimize a cost function. In this case, the system's hardware imposes physical limitations. The payload must remain in view of the camera on board for accurate state estimation. Operating on these constraints at the execution level, a MPC minimizes the error corresponding to a reference trajectory. Mathematically, the objective is the following constrained optimization problem
\begin{equation}
    \begin{split}
        \min_{\vecx{},\vecu{}}\int_{t_0}^{t_0+t_h} & L\left(\vecx{},\vecu{}\right)dt\label{eq:MPC_Form},\\ 
        \text{subject to:}\\
    \end{split}
\end{equation}
\begin{equation*}
    \begin{split}
    \centering
\dot{\vecx} &= \mathbf{f}(\vecx{},\vecu{}),~\forall t \in [t_0,t_0 + t_h],\\
\vecx{}_0 &= \vecx{}(t_0),\\
\mathbf{g}&(\vecx{},\vecu{})\leq0.
    \end{split}
\end{equation*}
In the equations above $\vecx{}$ and $\vecu{}$ are state and input vectors, respectively. The goal in mind is to obtain a series of inputs that optimize the objective function. The optimization occurs over a fixed look--ahead time horizon $\left[t_0,~t_0 + t_h\right]$ with initial condition $\vecx{}_0$, system dynamics $\mathbf{f}\left(\vecx{},\vecu{}\right)$ as well as the state and input constraints $\mathbf{g}\left(\vecx{},\vecu{}\right)$ to be satisfied. The error is determined by comparing predicted states and inputs to trajectory references. After a payload trajectory is generated, references are obtained by exploiting differential flatness as described in Section~\ref{sec:trajectory_planning}.
The formulation of eq.~(\ref{eq:MPC_Form}) is a nonlinear constrained optimization problem. In our work, Sequential Quadratic Programming (SQP) is used to approximate the model by iteratively solving a Quadratic Programming (QP) subproblem~\cite{Houska2011ACADOTO}. Furthermore, for accurate representation of the PCMPC, the optimization must occur within a desired frequency. Hence the model dynamics is discretized with a time step $dt$ for the fixed time horizon $t_h$ into $\vecx_i,~i = 0,\cdots,N$ and $\vecu_i,~i= 0,\cdots,N-1$. This results in a fixed horizon length $\left[0,N\right]$, where, $N = \frac{t_h}{dt}$, and $N$ is selected based on performance for the real-time implementation. The resulting discrete cost function is
\begin{equation}
    \min_{\vecx{},\vecu{}}  \frac{1}{2}\Tilde{\mathbf{x}}_N^\top \mathbf{Q}_x\Tilde{\mathbf{x}}_N + \sum_{i=0}^{N-1}\frac{1}{2}\Tilde{\mathbf{x}}_i^\top \mathbf{Q}_x\Tilde{\mathbf{x}}_i + \frac{1}{2}\Tilde{\mathbf{u}}_i^\top \mathbf{Q}_u\Tilde{\mathbf{u}}_i.
\end{equation}
The state and input error are $\Tilde{\mathbf{x}}_i=\mathbf{x}_{des,i} -\mathbf{x}_i$ and $\Tilde{\mathbf{u}}_i=\mathbf{u}_{des,i} -\mathbf{u}_i$ respectively. The desired values $\mathbf{x}_{des,i}$ and $\mathbf{u}_{des,i}$ are obtained exploiting the aforementioned differential flatness property as described in Section~\ref{sec:trajectory_planning}. The initial state $\mathbf{x}_0$ is obtained using visual inertial estimation as mentioned in Section~\ref{sec:state_estimation}. The constant cost diagonal matrices applied to the state and input are $\mathbf{Q}_x$ and $\mathbf{Q}_u$.
\subsubsection{Actuator and Dynamics Constraints}
Our system considers hardware limitations by constraining the control inputs
\begin{align}
    f_{min}\leq & f\leq f_{max}\label{eq:force_constraint},\\
    \Omega_{j,min}\leq&\Omega_{j}\leq\Omega_{j,max},\label{eq:angvel_constraint}
\end{align}
where $f_{min}$ and $f_{max}$ are the maximum and minimum thrusts, whereas $\Omega_{j,min}$ and $\Omega_{j,max}$ are the maximum and minimum angular velocity respectively, and $j={x,y,z}$.
Moreover, it is necessary to enforce the cable tautness during the system motion. To realize this, we first exploit the Newton equation of the forces on the system 
\begin{equation}
    -T\cablevec{} = m_L\left(\loadacc{}+g\vece{}^{\mathcal{I}}_z\right),
    \label{eq:system_forces}
\end{equation}
where $T = m_L\Vert\loadacc{}+g\vece{}^{\mathcal{I}}_z\Vert_2$ is the cable tension. To guarantee cable tautness, we need to guarantee the following constraint $T>0$. By projecting the right hand side of eq.~(\ref{eq:system_forces}) on the $\vece{}^{\mathcal{I}}_z$ axis, we notice that to ensure the cable tautness it is sufficient to ensure $\loadacc{}^\top\vece{}^{\mathcal{I}}_z>-g$. We guarantee that this condition is met when planning a trajectory as well.

\subsubsection{Perception and Sensing Constraints}
Our system depends on on--board sensing with perception--based feedback for payload state estimation. Therefore, on top of constraints eqs.~(\ref{eq:force_constraint}) and~(\ref{eq:angvel_constraint}), we apply a FOV constraint to guarantee the payload’s visibility during motion. In this way, the PCMPC can leverage the quadrotor's full agility while keeping the payload within the camera's FOV. In~\cite{JacquetICRA2020}, for the case of a quadrotor without a cable--suspended payload, the authors propose to incorporate the visibility of several image features within the camera's FOV. To achieve this goal it is sufficient to add a simple inequality constraint imposing each feature bearing be within the cone angle that represents the camera's FOV. In this work, the proposed system is more complex since the FOV constraint depends on the vehicle's and cable's motion. A similar approach to~\cite{JacquetICRA2020}, could work in the suspended payload case only if the camera frame $\mathcal{C}$ and robot frame $\mathcal{B}$ origins are coincident because of the cable swing motions. In case this assumption does not hold as shown in Fig.~\ref{fig:system_overview}, it is necessary to formulate and solve the FOV constraint problem in a more general way. 

We model the detection range of the camera as a conical shape with the apex at the origin of $\mathcal{C}$ with height, $h$, and radius, $r$. We refer to Fig.~\ref{fig:system_overview} for essential parameters that help define the perception constraint. The unit vector $\mathbf{n}$ with respect to $\worldf$ and magnitude $d$ describe the distance from the camera frame to the payload's center of mass (COM). We project $\mathbf{n}d$ onto $\vece{}^{\mathcal{C}}_z$ such that it aligns with the vertical axis of the FOV cone: $\mathbf{n}_{proj} = \prths{d\mathbf{n}^\top\vece{}^{\mathcal{C}}_z}\vece{}^{\mathcal{C}}_z$. The normal projection is at a point with radius $r_o$ perpendicular to $\vece{}^{\mathcal{C}}_z$ where it meets the bounds of the FOV: $r_o = \Vert\mathbf{n}_{proj}\Vert_2\frac{r}{h}$. Then, we calculate the orthogonal distance to the COM, $r_p$, on the plane made by $\vece{}^{\mathcal{C}}_x$ and $\vece{}^{\mathcal{C}}_y$: $r_p = \Vert\mathbf{n}d-\mathbf{n}_{proj}\Vert_2$. Finally, we define the FOV constraint such that $r_p$ must be less than or equal to $r_o$
\begin{equation}
    \norm{\mathbf{n}d -\mathbf{n}_{proj}}_2\leq
    \norm{\mathbf{n}_{proj}}_2\frac{r}{h}.
\end{equation}

Our approach also considers the acceleration and angular velocities ranges detectable by our IMU. Without loss of generality, let us assume that the IMU frame is coincident with the frame $\mathcal{B}$. The angular velocity constraint expressed by eq.~(\ref{eq:angvel_constraint}) incorporates the gyro limits on the vehicle as well. It is sufficient to select the most restrictive bounds among the commanded angular speed and the one that can be detected by the gyros. Furthermore, a similar constraint can be employed for the acceleration to guarantee that this fits within the detectable accelerometer bounds
\begin{equation}
     a_{j,min}\leq a_{j}\leq a_{j,max},
     \label{eq:acc_constraint}
\end{equation}
where $j={x,y,z}$. To impose this constraint over the state variables, it is then necessary to find the relationship between the vehicle's and the load's acceleration. In fact, the IMU provides the acceleration of the body frame and the system state includes the load position and its derivative instead. Considering, that
\begin{equation}
   \robotpos{Q} = \loadpos{}-l\cablevec{},
   \label{eq:quadrotor_position}
\end{equation}
and leveraging eq.~(\ref{eq:system_forces}), we obtain
\begin{equation}
   \robotpos{Q} =\loadpos{}+l\frac{\loadacc{}+g\vece{}^{\mathcal{I}}_z}{\Vert\loadacc{}{}+g\vece{}^{\mathcal{I}}_z\Vert_2}. \label{eq:cable_direction_geo}
\end{equation}
Taking the second order derivative of this expression, we notice that in eq.~(\ref{eq:acc_constraint}), $a_j=\ddot{\robotpos{}}_{Q} = g_{a_j}\prths{\loadpos{}^{(4)},\ddot{\mathbf{x}}_L}$.

\subsection{Trajectory Planning}\label{sec:trajectory_planning}
In our work, we plan trajectories exploiting the differential flatness of a quadrotor cable-suspended payload. The states and inputs in a differentially flat system  can be expressed by differential flat variables, $\vecx_f$, and a finite number of its derivatives. For instance, $\vecx{}$ and $\vecu{}$ can be expressed as functions: $\vecx{} = h_0(\vecx{}_f,\dot{\vecx{}}_f,\cdots\vecx{}_f^{\left(\alpha\right)}), ~\vecu{}= h_1(\vecx{}_f,\dot{\vecx{}}_f,\cdots\vecx{}_f^{\left(\alpha\right)})$, where $\alpha$ is an integer. This system uses flat variables  $\vecx{}_f = \left[\loadpos{}^\top, \psi\right]$. 
Eqs~(\ref{eq:acc})--(\ref{eq:rotmat}) and eq.~(\ref{eq:cable_direction_geo}) show that $\vecx{}$ and $\vecu{}$ can be written as a function of $\vecx{}_f$ and its derivatives. For all tests we implement smooth $\loadpos{}$ trajectories and assume the quadrotor heading angle, $\psi$, is constant. Thus, we guarantee a finite number of their derivatives exist and satisfy the conditions for differential flatness. The highest derivative order needed for the flat variables is $\loadpos{}^{(5)}$, $\dot{\psi}$. Lastly, to plan trajectories that produce a taut cable, we enforce the constraint in eq.~(\ref{eq:system_forces}) at the planning level as well.

\section{State Estimation}~\label{sec:state_estimation}
In this section, we introduce the perception and state estimation approaches for our system. As previously stated, one of the main goals in this work is to achieve closed-loop control of payload's position with on--board sensors. We consider using a minimal sensor suite, which consists of a single monocular camera and an IMU~\cite{LoiannoRAL2017}, for autonomous navigation. However, the sensing and perception problems of this system is more complex compared to the quadrotor case without suspended payloads. By inspecting eqs.~(\ref{eq:acc})-(\ref{eq:quatdot}), we observe that to address the vision--based control problem, we need to concurrently estimate the position and velocity of the payload as well as the cable direction $\cablevec{}$ and velocity $\cabledotvec{}$. To achieve this, we decouple the problem in two parts. First, we employ on--board Visual Inertial Odometry (VIO) to estimate the pose of the robot in the inertial frame $\mathcal{I}$. Second, the camera information is jointly used with motor speeds and VIO to estimate $\cablevec{}$ and velocity $\cabledotvec{}$ in the $\mathcal{I}$ frame.

\subsection{Visual Inertial Odometry}~\label{sec:vio}
The robot 6 Degrees of Freedom (DoF) position and orientation are estimated using a monocular downward facing camera and IMU available on the robot. The state estimation algorithm is based on a Unscented Kalman Filter-based VIO approach. The filter combines extracted image features with accelerometer and gyro data. The approach runs at the IMU rate. For more details on the localization pipeline, the reader can refer to our previous work~\cite{LoiannoRAL2017}.

 \begin{figure*}[!t]
  \centering
    \subfigure[]{\includegraphics[width=0.32\textwidth]{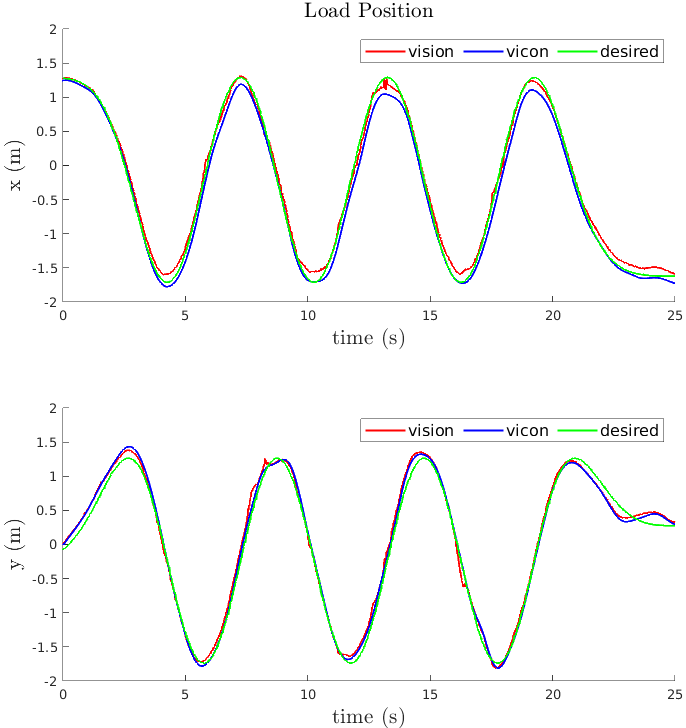}}
    \subfigure[]{\includegraphics[width=0.32\textwidth]{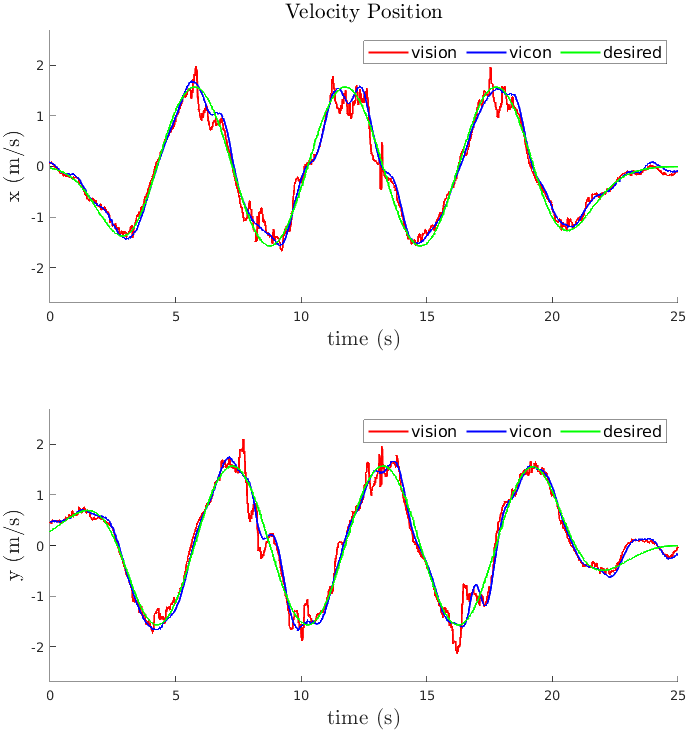}}
    \subfigure[]{\includegraphics[width=0.32\textwidth]{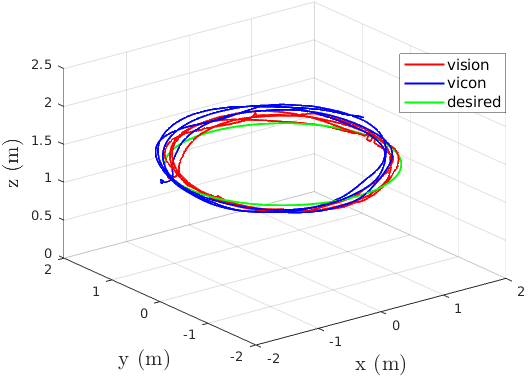}}
  \caption{Tracking results of the payload following a circle trajectory, (a) Cartesian position, (b) velocity, and (c) 3D path. \label{fig:circle_traj_tracking}}
\end{figure*}
\subsection{Cable State Estimation}\label{sec:cable_esitimation}
To estimate the cable direction, we employ the downward facing camera such to measure the bearing or cable direction. In this paper, an algorithm detecting black and white circular tag~\cite{whycon_jint} is employed to identify the cable attachment point on the payload and infer the cable direction. It provides fast, close to camera rate, and precise detection of a black and white roundel as shown in Fig.~\ref{fig:Inro}. From the algorithm, the pixel position $\begin{bmatrix}u&v\end{bmatrix}^{\top}$ of the circular tag in the image is obtained. We also want to point out that any other detector providing just the cable direction estimate can be used. The camera is modeled as a pinhole camera and the affine transformation between the pixel coordinates and the unit vector of tag position in $\qcamframe{}$ is 
\begin{equation*}
     \vecn^{\camf} = \begin{bmatrix}n_x\\n_y\\n_z\end{bmatrix}=\lambda\matK^{-1}\begin{bmatrix}u\\v\\1\end{bmatrix}, 
\end{equation*}
where $\lambda$ is a constant for the scale ambiguity, $\matK$ is the camera intrinsic matrix which we assume to be known from camera calibration. As shown in Fig.~\ref{fig:system_overview}, based on taut cable assumption, we have
\begin{equation}
    \aptpos{}^{\robotf} = \vecx_{\camf}^{\robotf} + d\mat{R}{\camf}^{\robotf}\vecn^{\camf},
\end{equation}
where $\aptpos{}^{\robotf},~\vecx_{\camf}^{\robotf}$ are the positions of the attach point and the camera with respect to $\robotf$ respectively, $\mat{R}{\camf}^{\robotf}$ is the rotation matrix of the frame $\camf$ with respect to the frame $\robotf$. After $\vecn^{\camf}$ is obtained, the depth $d$ of the tag in $\camf$ can be obtained by solving eq.~(\ref{eq:cable_constraint}) 
\begin{equation}
l = \norm{\vecx_{\camf}^{\robotf} + d\mat{R}{\camf}^{\robotf}\vecn^{\camf}}_2. \label{eq:cable_constraint}   
\end{equation}
The solution that corresponds to the attach point oriented below the robot is selected.
An Extended Kalman filter is designed to estimate the direction of the cable $\cablevec{}$ and its corresponding velocity $\cabledotvec{}$. The filter state and input are defined as
\begin{equation}
    \matX = \begin{bmatrix}\cablevec{}^{\top}&\cabledotvec{}^{\top}\end{bmatrix}^{\top},~
    \matU = \inputforce{},
\end{equation}
where $\inputforce{}$ is obtained based on measured motor speed values from the electronic speed controllers as
\begin{equation}
    \inputforce{} = \prths{\sum_{j=1}^{4}k_f\omega_{mj}^2}\matR\axis{3}{},\label{eq:motor_speed}
\end{equation}
with $k_f$ is the known motor constant, $\omega_{mj}$ is the motor speed of the $j^{th}$ motor. Based on the dynamics of the cable in eq.~(\ref{eq:cable_acc}), we can obtain the resulting process model as

\begin{equation}
\begin{split}
\dot{\matX} = \begin{bmatrix}\cabledotvec{}\\\frac{1}{m_Q l_{}}\cablevec{}\times\prths{\cablevec{}\times\inputforce{}}-\twonorm{\cabledotvec{}}^2\cablevec{}\end{bmatrix}+ \mat{N}{}
\end{split}
\end{equation}
where the process noise $\mat{N}{}\in\realnum{6}$ is modeled as additive Gaussian white noise $\mat{N}{}\sim\mathcal{N}(\mathbf{0},Q_{\mat{N}{}})$ with zero mean and covariance $Q_{\mat{N}{}}\in\realnum{6\times6}$. 
The measurement model is 
\begin{equation}
\begin{split}
\mat{Z}{} &= \begin{bmatrix}\aptpos{}^{\robotf}\\\vspace{-8pt}\\\aptvel{}^{\robotf}\end{bmatrix} =  g\prths{\mat{X}{},\mat{V}{}}= \begin{bmatrix}\matR^{\top}l\cablevec{}\\l\prths{\matR^{\top} \cabledotvec{}-\robotangvel{}\times\matR^{\top}\cablevec{}}\end{bmatrix} + \mat{V}{},\label{eq:cable_measurement_model}
\end{split}
\end{equation}
where the measurement noise is modeled as additive Gaussian white noise $\mat{V}{}\sim\mathcal{N}(\mathbf{0},Q_{\mat{V}{}})$ with zero mean and standard deviation $Q_{\mat{V}{}}\in\realnum{6\times 6}$.

\section{Experimental Results}~\label{sec:experimental_results}

The experiments are performed in an indoor testbed with a flying space of $7\times5\times4~\si{m^3}$ at the ARPL lab at the New York University. The ground truth data is collected using a Vicon\footnote{\url{www.vicon.com}} motion capture system at $100~\si{Hz}$. The mass of the payload is $75~\si{g}$ and the cable length is $0.5~\si{m}$. The $250~\si{g}$ quadrotor platform used in the experiments is equipped with a $\text{Qualcomm}^{\circledR}\text{Snapdragon}^{\text{TM}}$ board for on--board computation~\cite{LoiannoRAL2017}.  The control and estimation frameworks are developed in ROS\footnote{\url{www.ros.org}}.
Our approach runs on--board at $500$ Hz both for control and state estimation and using $90\%$ of the overall CPU. For the PCMPC controller, we choose the prediction time horizon $t_h$ as $1~\si{s}$, the discretized time step $dt$ as $0.1~\si{s}$ and the horizon length $N = 10$. It has been implemented in ACADO~\cite{Houska2011ACADOTO} using the qpOASES~\cite{qpOASES} solver.

\begin{table}[!t]
\caption {Payload trajectory tracking error and RMSE at different maximum speed.\label{tab:tracking_error}} 
\centering
\begin{tabularx}{0.48\textwidth}{bbcccc}
    \hline\hline
 \rule{0pt}{2ex} &Statistics&Component&$T_c =9s$&$T_c =6s$&$T_c =4s$\\\hline
  Position    &MEAN &x& 0,06913 & 0.06617 & 0.1801  \\
  $(\si{m})$  &ERROR&y& 0.09920 & 0.08698 & 0.2306  \\
              &     &z& 0.03286 & 0.05383 & 0.05741 \\\cline{2-6}
              &RMSE &x& 0.08038 & 0.08076 & 0.2143  \\
              &     &y& 0.1168  & 0.1062  & 0.2627  \\
              &     &z& 0.03678 & 0.06130 & 0.07153 \\\hline
  Velocity    &MEAN &x& 0.1039  & 0.1400  & 0.2030  \\  
  $(\si{m/s})$&ERROR&y& 0.1099  & 0.1503  & 0.1485   \\
              &     &z& 0.05345 & 0.06232 &  0.08574 \\\cline{2-6}
              &RMSE &x& 0.1337  & 0.1898  &  0.2621  \\      
              &     &y& 0.1319  & 0.2070  &  0.1895  \\      
              &     &z& 0.06799 & 0.07972 &  0.1096 \\\cline{2-6}
 \hspace{1pt} &Max Speed&\hspace{1pt} &\multirow{2}{3em}{1.20}&\multirow{2}{3em}{1.86}&\multirow{2}{3em}{2.47}\\\hline
    \hline                     
\end{tabularx}                 
\end{table}
\begin{table*}[!t]
\caption {Mean and STD of absolute error and RSME of visual payload state estimation during circular trajectory.\label{tab:pose_estimation_error}} 
\centering
\begin{tabularx}{0.9\textwidth}{bbccccccc}
    \hline\hline
  Circular Trajectory Period&Statistics&\multicolumn{3}{c}{Position ($\si{m}$)}&&\multicolumn{3}{c}{Velocity ($\si{m/s}$)}\\
  \cline{3-5}\cline{7-9}
            &      & x    & y &  z & &  x  & y &  z\\\hline
    $T_c =9s$&MEAN& 0.04150 & 0.1324 & 0.05546&&0.08427 & 0.07019 & 0.04606\\
             &STD& 0.02673 & 0.04484 & 0.04590&&0.06721 & 0.05496 & 0.03386\\
             &RMSE&0.04936 & 0.13980 & 0.07199&&0.10779 & 0.08915 & 0.05717\\\hline
    $T_c =6s$&MEAN& 0.1294 & 0.04400 & 0.08518&&0.1070 & 0.1110 & 0.04981 \\
          &STD& 0.04805 & 0.03626 & 0.06582&&0.1028 & 0.1199 & 0.03911\\
          &RMSE&0.1381 & 0.05702 & 0.1076&&0.1484 & 0.1634 & 0.06333\\\hline
    $T_c =4s$&MEAN& 0.07779 & 0.08548 & 0.05195&&0.1449 & 0.1275 & 0.05678 \\
          &STD& 0.04243 & 0.04749 & 0.04337&&0.1162 & 0.1057 & 0.04489\\
          &RMSE&0.08861 & 0.09778 & 0.06767&&0.1857 & 0.1656 & 0.07238\\
    \hline\hline
\end{tabularx}
\end{table*}
\begin{table}[!t]
\caption {Mean and STD of the cable estimation errors.\label{tab:cable_estimation}} 
\centering
\begin{tabularx}{0.48\textwidth}{bbbb}
    \hline\hline
 \rule{0pt}{2.5ex}Trajectory&Statistics&$e_{angle} ~(\si{rad})$&$e_{\cabledotvec{}}~(\si{1/s})$\\\hline
$T_c =9~\si{s}$&MEAN&0.03278 & 0.1701\\
               &STD& 0.01399 & 0.08951\\\hline
$T_c =6~\si{s}$&MEAN&0.05920 & 0.1993\\
               &STD& 0.01985 & 0.1635\\\hline
$T_c =4~\si{s}$&MEAN&0.02224 & 0.1944\\
               &STD& 0.01497 & 0.1197\\
    \hline\hline
\end{tabularx}
\end{table}

\subsection{Payload Trajectory Tracking}
We test our approach considering a payload's trajectory tracking problem for a circular trajectory
\begin{equation}
\loadposdes(t) = \begin{bmatrix}r\cos\frac{2\pi t}{T_c}&r\sin\frac{2\pi t}{T_c}&h_c\end{bmatrix}^\top
\end{equation}
 with period $T_c = 9, 6, 4~\si{s}$, radius $r = 1.5~\si{m}$. The other states in the state vector are derived using the differential flatness property of the system. We used the on--board camera and IMU to make state estimations and compare these with the ground truth obtained by Vicon. We include Fig.~\ref{fig:circle_traj_tracking}, where $T_c = 6~\si{s}$, as a sample plot of our payload tracking experiment. The other experiments have similar tracking performance. Table.~\ref{tab:tracking_error} shows consistent tracking behavior across circular trajectory favoring the position over velocity tracking. This pattern was expected since we designed a cost function that has inherent preference for position over the velocity. In the attached multimedia material, we provide several additional tests with different trajectories and similar results to the one reported in this section and payload's velocities up to $\sim 4~\si{m/s}$  with similar performances to the results reported in this paper.

\subsection{State Estimation}
\begin{figure}[!t]
  \centering
  \includegraphics[width=1\columnwidth]{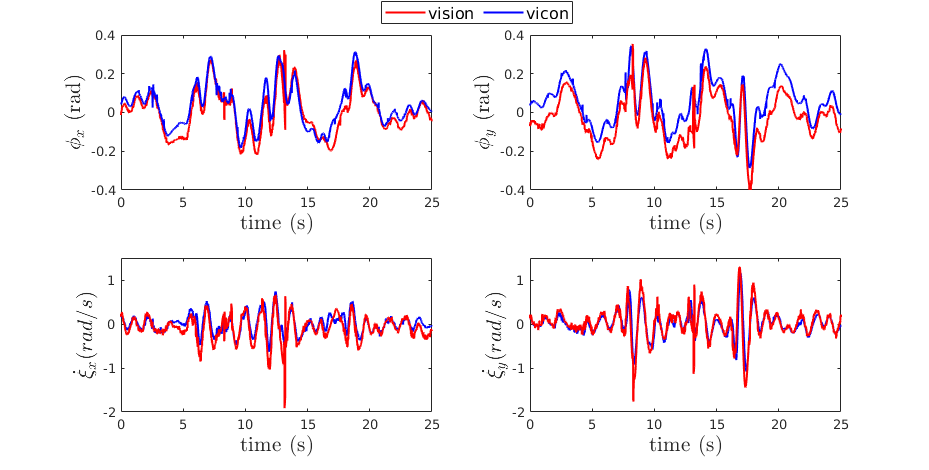}
  \caption{Estimation results of the cable direction and velocity. \label{fig:lissajous_dgf2_cable_direction_velocity}}
  \vspace{-10pt}
\end{figure}

We evaluate the visual estimation of the cable state using the angle errors to evaluate the estimation of cable direction. The errors were determined by comparing vision to ground truth which are denoted with $(\cdot)_v$, $(\cdot)_{gt}$, respectively. The angle is calculated with the cable direction $\cablevec{}=\brcks{\xi_x, \xi_y,\xi_z}^\top$
 \begin{equation}
     e_{angle} = \frac{\abs{\phi_{x,gt}-\phi_{x,v}}+\abs{\phi_{y,gt}-\phi_{y,v}}}{2},
 \end{equation}
where
 \begin{equation}
     \phi_x = \tan^{-1}\left(\frac{\xi_{y}}{-\xi_{z}}\right),~\phi_y = -\tan^{-1}\left(\frac{\xi_{x}}{-\xi_{z}}\right).
 \end{equation}
We use $e_{\cabledotvec{}}$ to evaluate the estimation of the cable velocity
  \begin{equation}
     e_{\cabledotvec{}} = \twonorm{\cabledotvec{gt} - \cabledotvec{v}}.
 \end{equation}
In Fig.~\ref{fig:lissajous_dgf2_cable_direction_velocity} we show the on--board cable state estimation during the circular trajectory. The mean and Standard Deviation (STD) of the errors of cable direction and velocity estimation during the flight trajectory are shown in the Table~\ref{tab:cable_estimation}.
We also report the performance of the visual estimation method, which infers the payload's position and velocity. The estimation results are reported in Table~\ref{tab:pose_estimation_error} which shows the RMSE of both position and velocity tracking of the payload. Both tables show that small estimation errors of the payload's states and cable's states within few centimeters. Moreover, the payload remains in the camera FOV satisfying the perception constraint. 
\section{Conclusion}~\label{sec:conclusion}
In this paper, we presented the first MPC for quadrotors with cable--suspended payloads that incorporates perception, system dynamics, and actuators constraints and that is exclusively based on on--board perception and estimation. Our method runs in real--time on a small computational unit. Several trajectory tracking experiments with payload's velocity reaching up to $\sim 4$ m/s have shown the robustness of the proposed approach.

Future works will involve the ability to estimate the payloads' geometric and inertial characteristics using on--board data. We would like to extend this approach to multiple vehicles cooperatively transporting a payload as well as develop obstacle avoidance strategies based on on--board sensing. Finally, we would also like to extend the PCMPC considering a data--driven formulation to improve the transportation task speed and tracking performance over time based on the collected data during the transportation.

\addtolength{\textheight}{-9.0cm}   


\bibliographystyle{IEEEtran}	
\bibliography{references}
\end{document}